\documentclass{article}

\usepackage[final]{neurips_2022}

\usepackage{algorithm}
\usepackage{amsmath}
\usepackage{graphicx}
\usepackage{subcaption}
\usepackage{multirow}
\usepackage[utf8]{inputenc} 
\usepackage[T1]{fontenc}    
\usepackage{url}
\usepackage{booktabs}       
\usepackage{amsfonts}       
\usepackage{nicefrac}       
\usepackage{microtype}      
\usepackage{xcolor}         
\usepackage{graphicx}
\usepackage{rotating}
\usepackage[breaklinks=true]{hyperref}
\title{Exploring Multiple Neighborhood Neural Cellular Automata (MNNCA) for Enhanced Texture Learning}

\author{%
  Magnus ~Petersen \\
  FIAS\\
  Goethe University Frankfurt\\
  Ruth-Moufang-Straße 1, 60438 Frankfurt am Main \\
  \texttt{mapetersen@fias.uni-frankfurt.de} \\
}

\begin{document}

\maketitle

\begin{abstract}
Cellular Automata (CA) have long been foundational in simulating dynamical systems computationally. With recent innovations, this model class has been brought into the realm of deep learning by parameterizing the CA's update rule using an artificial neural network, termed Neural Cellular Automata (NCA). This allows NCAs to be trained via gradient descent, enabling them to evolve into specific shapes, generate textures, and mimic behaviors such as swarming. However, a limitation of traditional NCAs is their inability to exhibit sufficiently complex behaviors, restricting their potential in creative and modeling tasks. Our research explores enhancing the NCA framework by incorporating multiple neighborhoods and introducing structured noise for seed states. This approach is inspired by techniques that have historically amplified the expressiveness of classical continuous CA. All code and example videos are publicly available on \href{https://github.com/MagnusPetersen/MNNCA}{Github}.
\end{abstract}
\section{Introduction}
Cellular Automata (CA) are mathematical models composed of a grid of cells of either continuous or discrete states. The state of each cell evolves over time based on a set of rules that consider the states of neighboring cells. Historically, John von Neumann first conceptualized CA in the 1950s with the aim of creating a self-replicating system \cite{vonNeumann1966selfreproducing}. Later, in the 1970s, John Conway popularized CA with his "Game of Life,"\cite{conway1970gameoflife} a simple yet powerful demonstration of how basic rules can lead to complex and unpredictable patterns.

These models have since been recognized as a potent tool in the computational modeling of dynamical systems. Originating from simple rules and interactions, CA can produce intricate patterns and behaviors, making them an attractive model for a myriad of applications ranging from physics to biology. With the advent of deep learning, there has been a growing interest in merging traditional CA principles with neural networks, leading to the development of Neural Cellular Automata (NCA) \cite{gilpin2019cellular}. By parameterizing the update rules of CA with artificial neural networks, NCAs can be trained, adapted, and refined using gradient descent, opening up new avenues for modeling and simulation. This approach has been used to train the NCAs to grow into shapes \cite{randazzo2023growing} form textures \cite{mordvintsev2021texture} and has even been extended to graphs \cite{grattarola2021learning} to allow the modeling of an even wider range of dynamics.

However, while NCAs offer promising capabilities, they are not without limitations. Their potential in emulating complex behaviors is often limited. Recognizing this gap, our study seeks to augment the expressiveness of NCAs. Drawing inspiration from techniques developed by Slackermanz for continuous CA \cite{slackermanz2023understanding} that have enhanced CAs capabilities, we delve into the integration of multiple neighborhoods and the introduction of structured noise as NCA seed state leading to more complex and creatively interesting dynamics.
\section{Method}
Our approach to enhancing Neural Cellular Automata (NCA) involves two primary modifications. First, we integrate multiple neighborhoods into the NCA framework, allowing for richer interactions and dependencies between cells and multiscale dynamics if different neighborhood sizes are used. Second, we introduce structured noise, like Perlin noise \cite{perlin2002improving}, as the initial condition making the generation of structured textures easier for the automata. To test the improvement in complex behavior we trained the model on a texture loss \cite{gatys2016image}\cite{heitz2021sliced}.
\begin{figure}[htbp]
    \centering
    \includegraphics[width=0.95\textwidth]{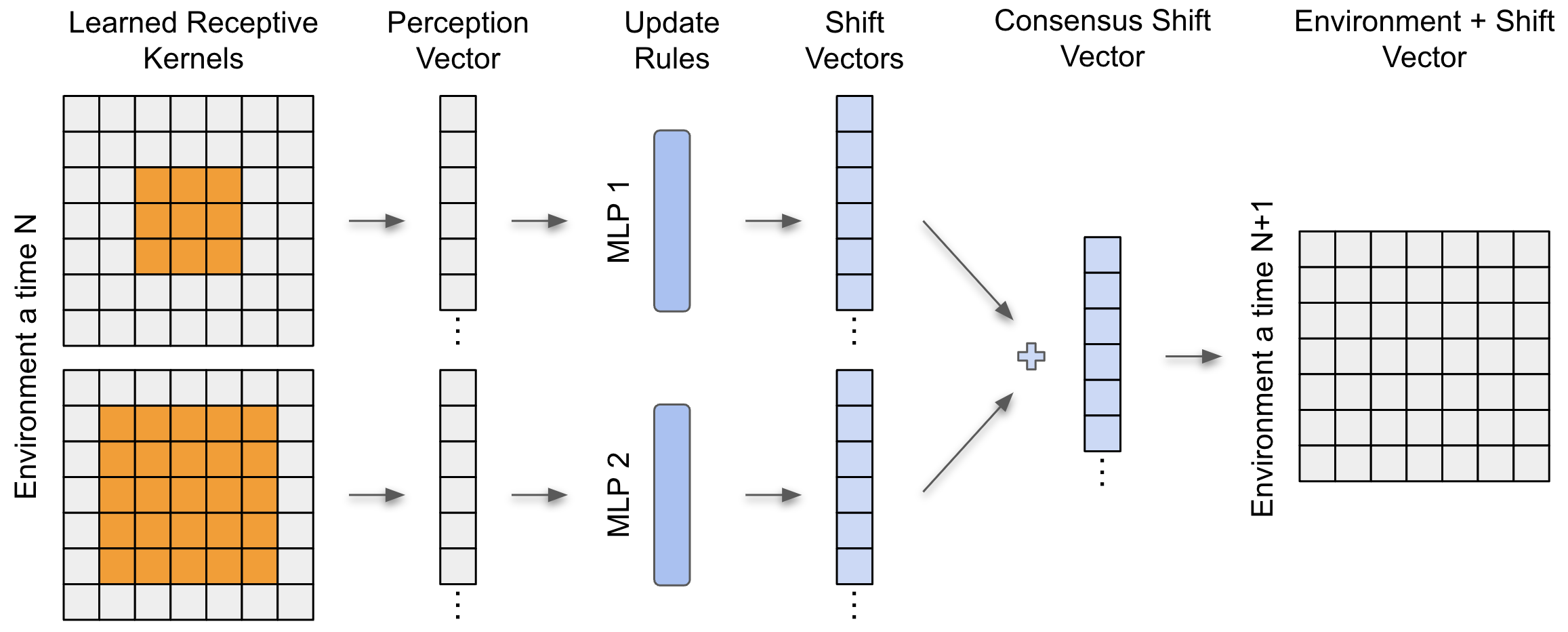}
    \caption{Illustration of the MNNCA algorithm.}
    \label{fig:MNNCA_algorithm}
\end{figure}
\section{Results}
\begin{figure}[htbp]
    \centering
    \begin{subfigure}{0.32\textwidth}
        \includegraphics[width=\textwidth]{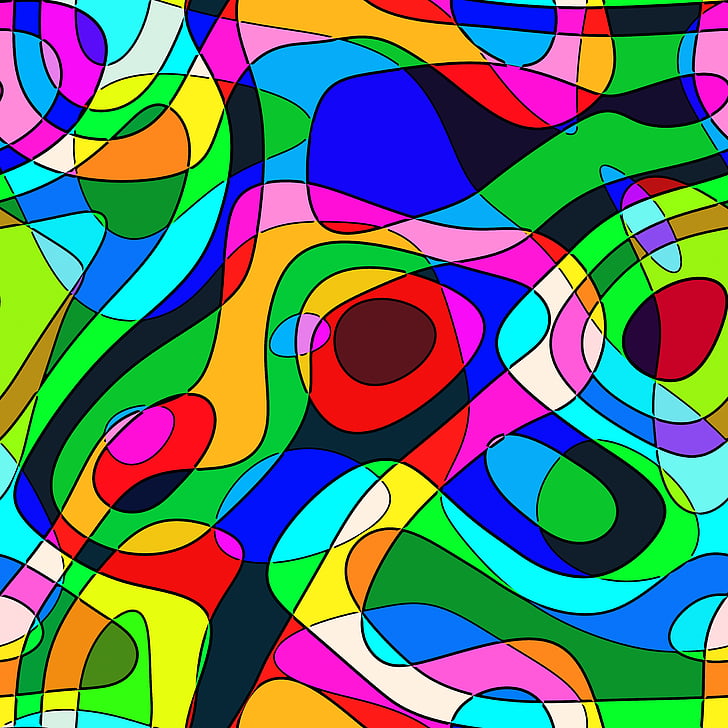}
        \caption*{(a) Target Texture \cite{hippopx2023image}}
    \end{subfigure}
    \begin{subfigure}{0.32\textwidth}
        \includegraphics[width=\textwidth]{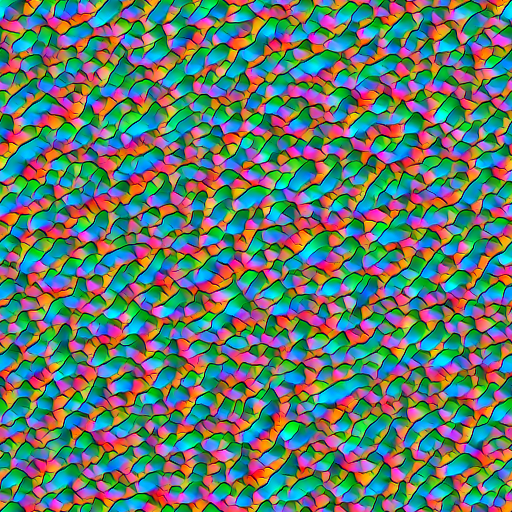}
        \caption*{(b) NCA after 50 time steps}
    \end{subfigure}
    \begin{subfigure}{0.32\textwidth}
        \includegraphics[width=\textwidth]{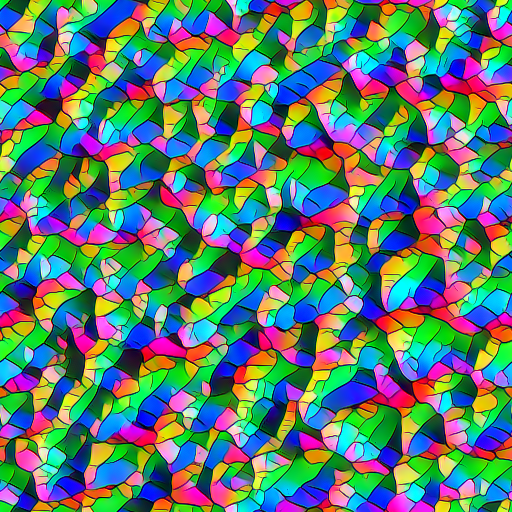}
        \caption*{(c) NCA after 200 time steps}
    \end{subfigure}
    
    \begin{subfigure}{0.32\textwidth}
        \includegraphics[width=\textwidth]{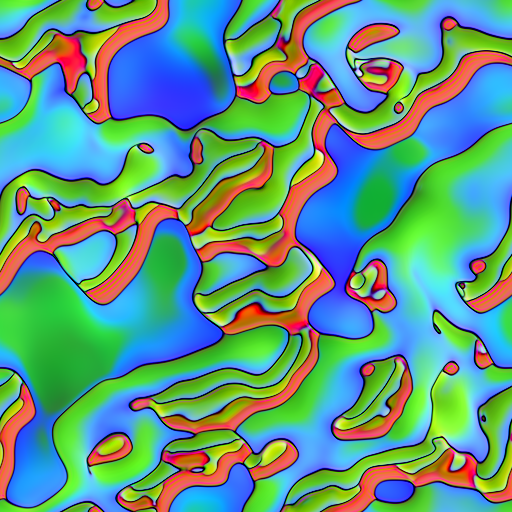}
        \caption*{(d) MNNCA after 50 time steps}
    \end{subfigure}
    \begin{subfigure}{0.32\textwidth}
        \includegraphics[width=\textwidth]{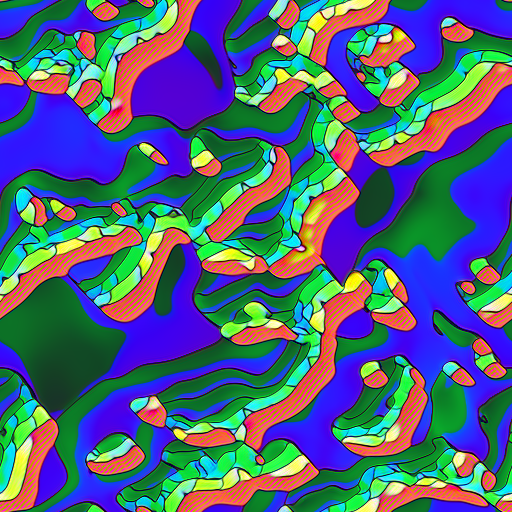}
        \caption*{(e) MNNCA after 200 time steps}
    \end{subfigure}
    \begin{subfigure}{0.32\textwidth}
        \includegraphics[width=\textwidth]{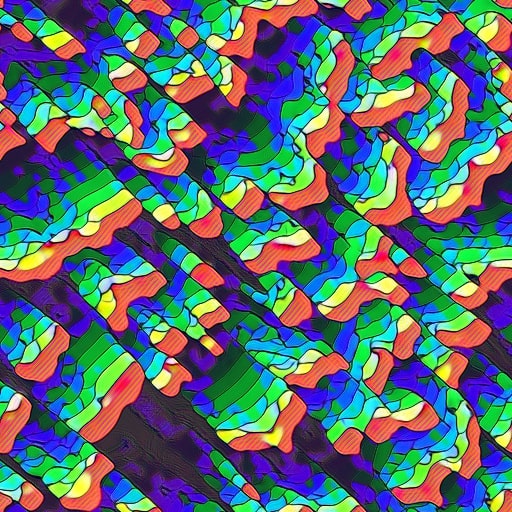}
        \caption*{(f) MNNCA after 600 time steps}
    \end{subfigure}
    
    \caption{Comparison of textures generated by NCA and MNNCA at different time steps. Both models were trained with a 256x256 resolution and sampled at 512x512.}
    \label{fig:texture_comparison}
\end{figure}
\newpage
\section{Ethical Implications}
Given the nature of our study, which primarily focuses on computational modeling and its technical advancements for topics like texture generation, we have not identified any direct ethical issues or implications. However, we acknowledge the importance of ongoing ethical reflection as this technology evolves and is applied in broader contexts.
\medskip
{
\small
\bibliography{bib}
}
\newpage
\section{Supplementary}
\section{Update Rules Explanation}

Our method can be used in different ways:

\begin{enumerate}
    \item \textbf{Sum Update}: This rule directly aggregates all MNNCA output channel values without any weighting or selection. Mathematically, it is represented as:
    \[
    y = \sum_{i} y_{\text{MNNCA output channel}[i]}
    \]

    \item \textbf{Random Update}: This rule employs a stochastic approach by generating a random mask to weight and mix the MNNCA output channel values. The mask is normalized using a softmax function to ensure the sum of weights equals one. The output is given by:
    \[
    y = \sum_{i} y_{\text{MNNCA output channel}[i]} \times \text{choice\_mask}[i]
    \]

    \item \textbf{Based on Environment Channel Value}: This rule derives weights from one of the channels of the environment to mix the MNNCA output channel values. For two outputs, the rule can be represented as:
    \[
    y = y_{\text{MNNCA output channel}[0]} \times \text{choice\_mask} + y_{\text{MNNCA output channel}[1]} \times (1 - \text{choice\_mask})
    \]
    For more than two outputs, the weights are derived from multiple channels of the environment channel value and normalized using a softmax function.

    \item \textbf{Based on MNNCA Output Channel Value}: This rule uses the MNNCA output channel values themselves to determine the weights for mixing. For two outputs, the rule is:
    \[
    y = y_{\text{MNNCA output channel}[0]} \times \text{choice\_mask} + y_{\text{MNNCA output channel}[1]} \times (1 - \text{choice\_mask})
    \]
    For multiple outputs, specific channels from all MNNCA output channel values are aggregated to determine the weights, which are then normalized using a softmax function.
\end{enumerate}
The demonstrations in the paper are based on 3. However, the code offers the ability to experiment with all four.
\subsubsection{Experimental Details}\label{sec:ExperimentalDetails}
\begin{table}[H]
  \caption{Experimental Settings}
  \label{tab:experimental_settings}
  \centering
  \begin{tabular}{llll}
    \toprule
    Model & Parameter Count & Rule Count & Batch Count \\
    \midrule
    NCA & 2806 & 1 & 3000 \\
    NCA XL & 37576 & 1 & 3000 \\
    MNNCA & 37536 & 2 & 3000 \\
    \bottomrule
  \end{tabular}
\end{table}
\newpage
\subsubsection{Ablation Experiments}
\begin{figure}[htbp]
    \centering
    
    \begin{sideways} NCA \& Uniform Noise\end{sideways}\quad
    \begin{subfigure}{0.31\textwidth}
        \includegraphics[width=\textwidth]{Figures/nca_basic_image_at_step_50.png}
    \end{subfigure}
    \begin{subfigure}{0.31\textwidth}
        \includegraphics[width=\textwidth]{Figures/nca_basic_image_at_step_200.png}
    \end{subfigure}
    \begin{subfigure}{0.31\textwidth}
        \includegraphics[width=\textwidth]{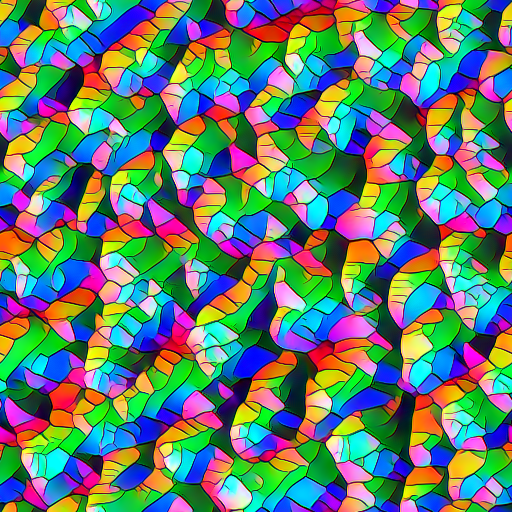}
    \end{subfigure}
    
    \begin{sideways} NCA \& Perlin Noise\end{sideways}\quad
    \begin{subfigure}{0.31\textwidth}
        \includegraphics[width=\textwidth]{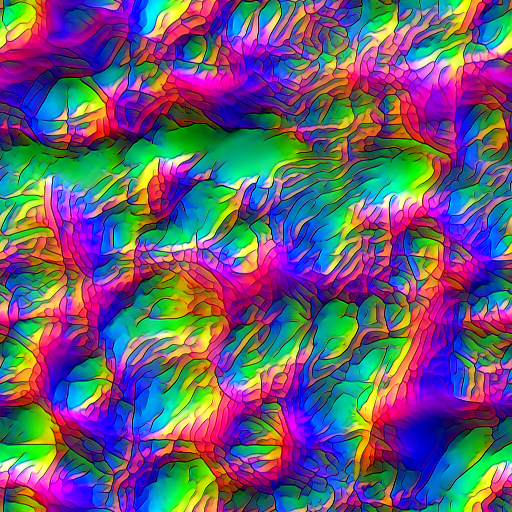}
    \end{subfigure}
    \begin{subfigure}{0.31\textwidth}
        \includegraphics[width=\textwidth]{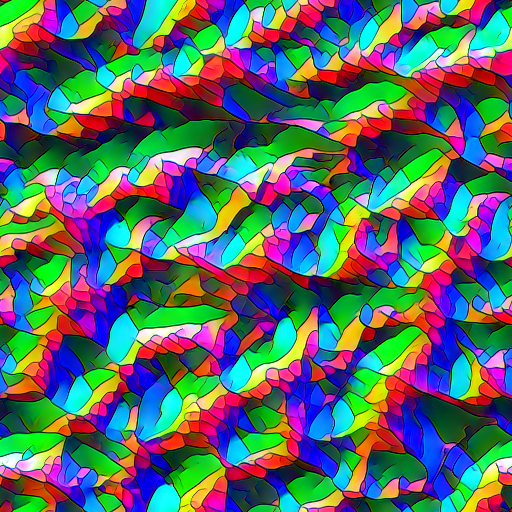}
    \end{subfigure}
    \begin{subfigure}{0.31\textwidth}
        \includegraphics[width=\textwidth]{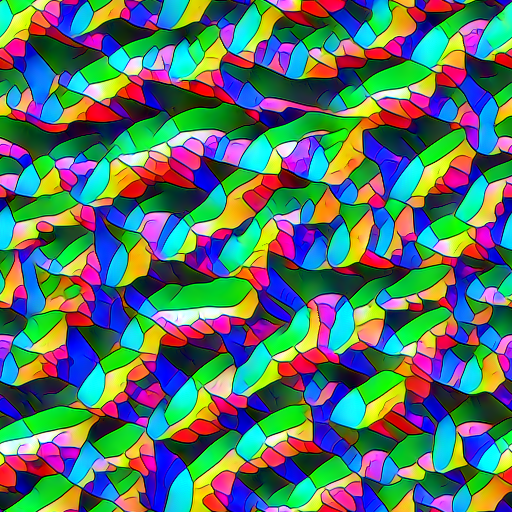}
    \end{subfigure}
    
    \begin{sideways} MNNCA \& Uniform Noise\end{sideways}\quad
    \begin{subfigure}{0.31\textwidth}
        \includegraphics[width=\textwidth]{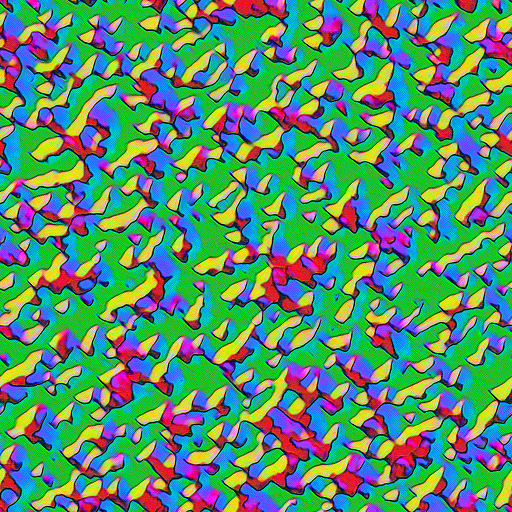}
    \end{subfigure}
    \begin{subfigure}{0.31\textwidth}
        \includegraphics[width=\textwidth]{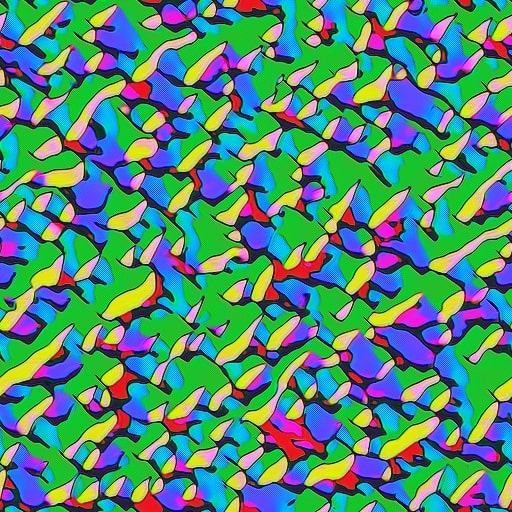}
    \end{subfigure}
    \begin{subfigure}{0.31\textwidth}
        \includegraphics[width=\textwidth]{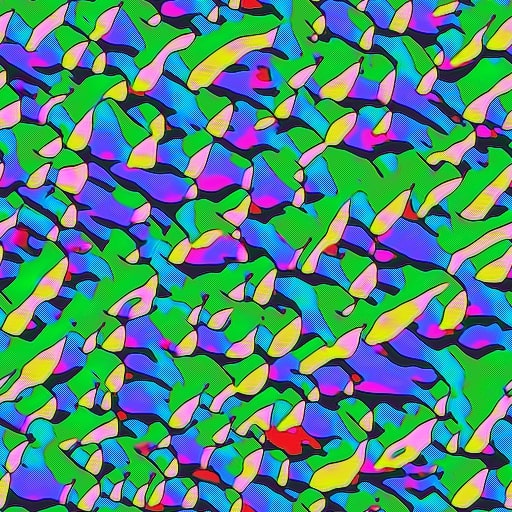}
    \end{subfigure}
    
    \begin{sideways} MNNCA \& Perlin Noise\end{sideways}\quad
    \begin{subfigure}{0.31\textwidth}
        \includegraphics[width=\textwidth]{Figures/mnnca_perlin_image_at_step_50.png}
    \end{subfigure}
    \begin{subfigure}{0.31\textwidth}
        \includegraphics[width=\textwidth]{Figures/mnnca_perlin_image_at_step_200.png}
    \end{subfigure}
    \begin{subfigure}{0.31\textwidth}
        \includegraphics[width=\textwidth]{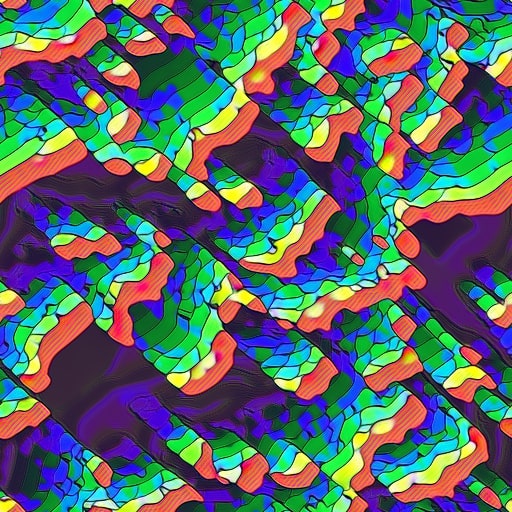}
    \end{subfigure}
    
    \caption{Illustrations of various timesteps, 50, 200, and 600, of classical NCA and MNNCA, with uniform and Perlin noise.\href{https://i0.hippopx.com/photos/521/378/436/digital-multicolor-colorful-curves-preview.jpg}{Trained on texture} \cite{hippopx2023image}.}
    \label{fig:algorithm_stages}
\end{figure}
\begin{figure}[htbp]
    \centering
    
    \begin{sideways} NCA \& Uniform Noise\end{sideways}\quad
    \begin{subfigure}{0.31\textwidth}
        \includegraphics[width=\textwidth]{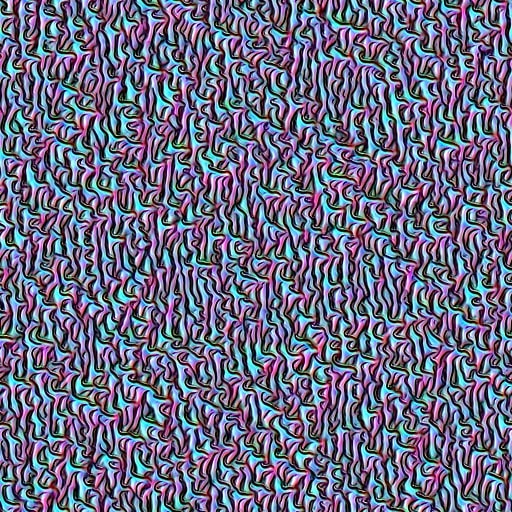}
    \end{subfigure}
    \begin{subfigure}{0.31\textwidth}
        \includegraphics[width=\textwidth]{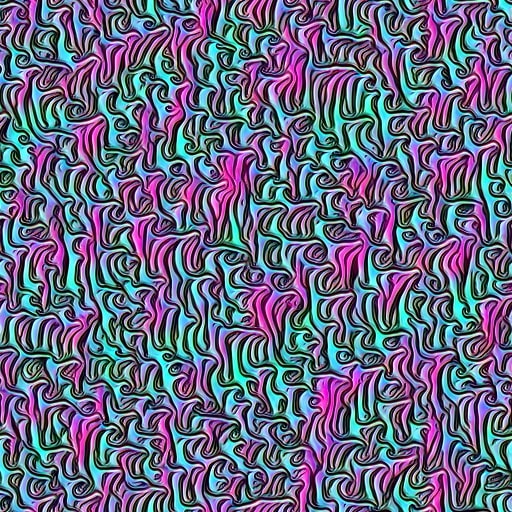}
    \end{subfigure}
    \begin{subfigure}{0.31\textwidth}
        \includegraphics[width=\textwidth]{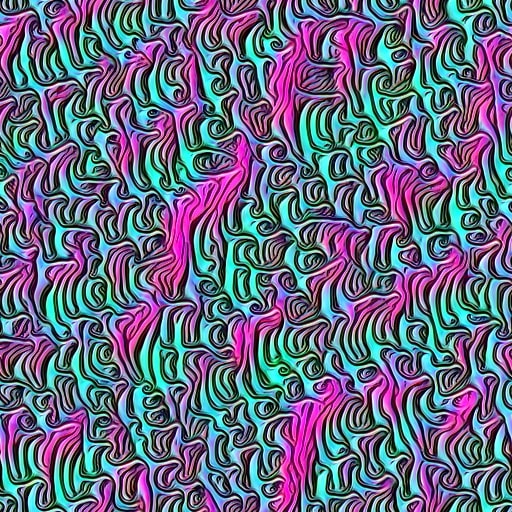}
    \end{subfigure}
    
    \begin{sideways} NCA \& Perlin Noise\end{sideways}\quad
    \begin{subfigure}{0.31\textwidth}
        \includegraphics[width=\textwidth]{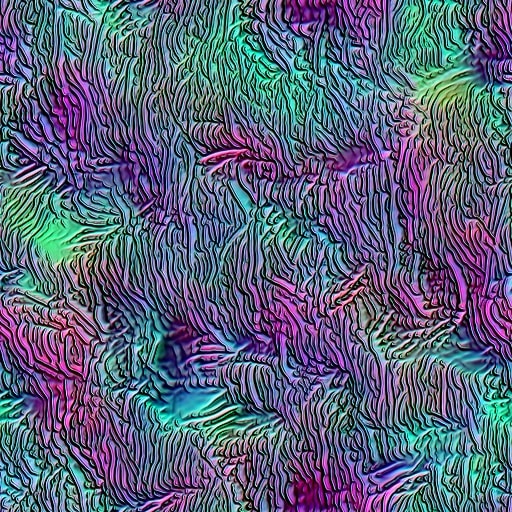}
    \end{subfigure}
    \begin{subfigure}{0.31\textwidth}
        \includegraphics[width=\textwidth]{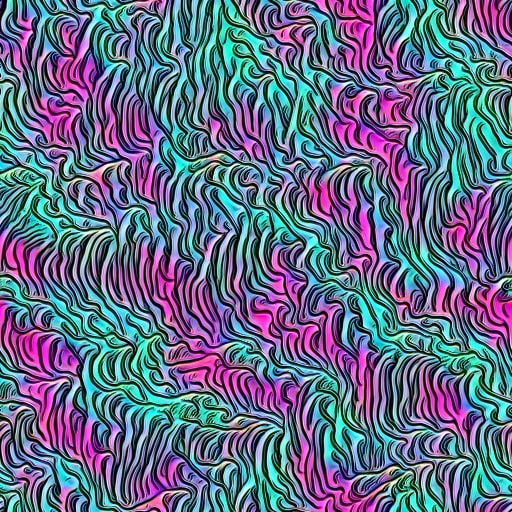}
    \end{subfigure}
    \begin{subfigure}{0.31\textwidth}
        \includegraphics[width=\textwidth]{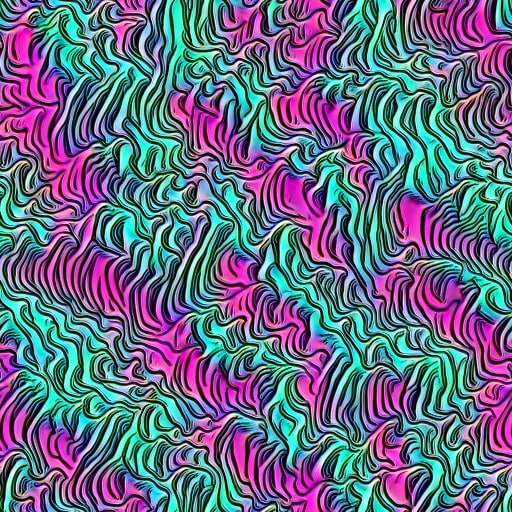}
    \end{subfigure}
    
    \begin{sideways} MNNCA \& Uniform Noise\end{sideways}\quad
    \begin{subfigure}{0.31\textwidth}
        \includegraphics[width=\textwidth]{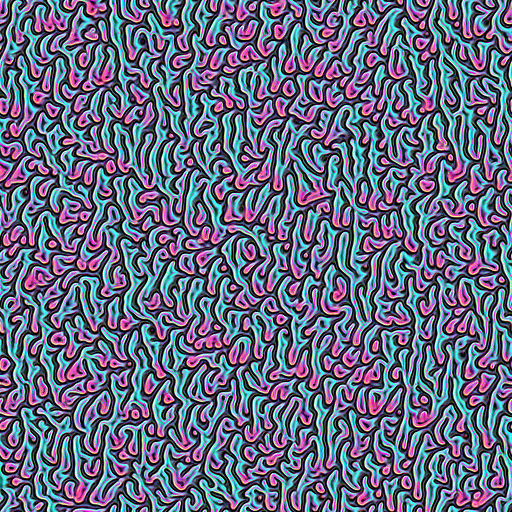}
    \end{subfigure}
    \begin{subfigure}{0.31\textwidth}
        \includegraphics[width=\textwidth]{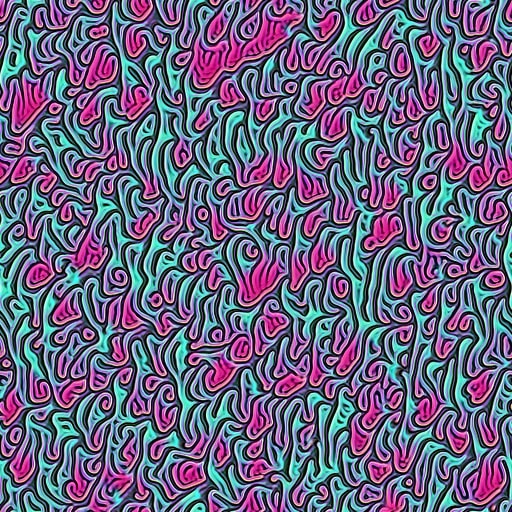}
    \end{subfigure}
    \begin{subfigure}{0.31\textwidth}
        \includegraphics[width=\textwidth]{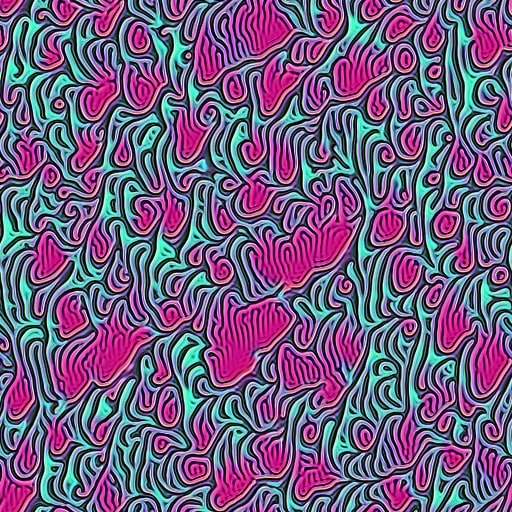}
    \end{subfigure}
    
    \begin{sideways} MNNCA \& Perlin Noise\end{sideways}\quad
    \begin{subfigure}{0.31\textwidth}
        \includegraphics[width=\textwidth]{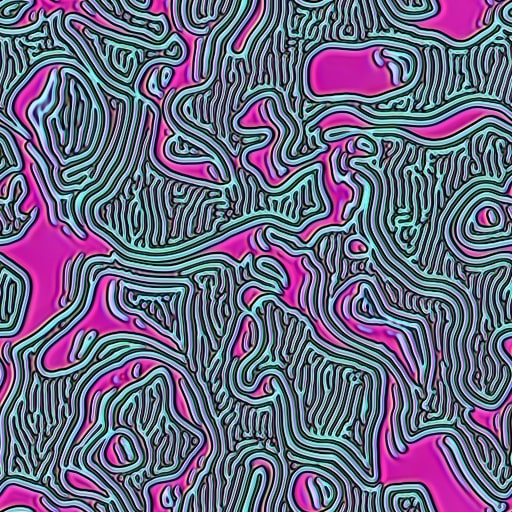}
    \end{subfigure}
    \begin{subfigure}{0.31\textwidth}
        \includegraphics[width=\textwidth]{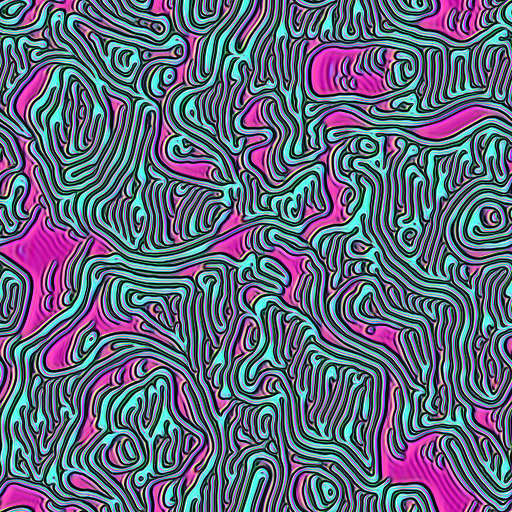}
    \end{subfigure}
    \begin{subfigure}{0.31\textwidth}
        \includegraphics[width=\textwidth]{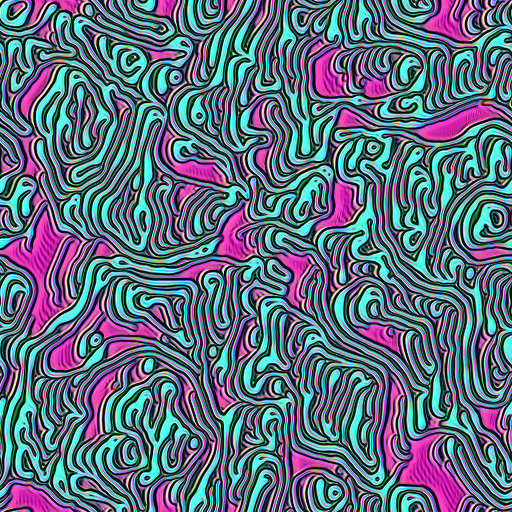}
    \end{subfigure}
    
    \caption{Illustrations of various timesteps, 50, 200, and 600, of classical NCA and MNNCA, with uniform and Perlin noise. \href{https://thumbs.dreamstime.com/b/trama-verniciata-con-arcobaleno-senza-saldatura-neretto-psichedelico-neon-artistico-scarabocchio-di-matite-lavato-imperfetto-190793728.jpg}{Trained on texture} \cite{dreamstimeimage}.}
    \label{fig:algorithm_stages}
\end{figure}
\begin{figure}[htbp]
    \centering
    
    \begin{sideways} NCA \& Uniform Noise\end{sideways}\quad
    \begin{subfigure}{0.31\textwidth}
        \includegraphics[width=\textwidth]{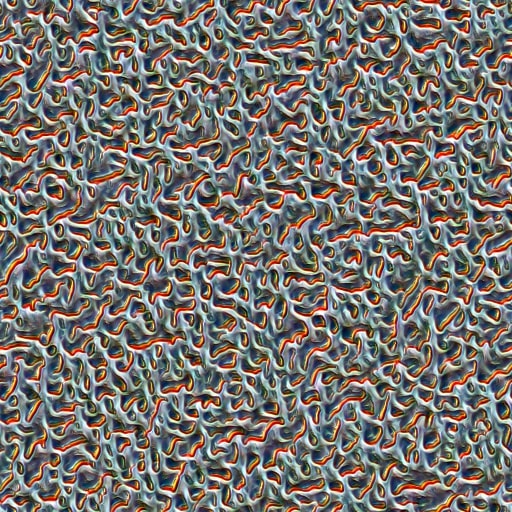}
    \end{subfigure}
    \begin{subfigure}{0.31\textwidth}
        \includegraphics[width=\textwidth]{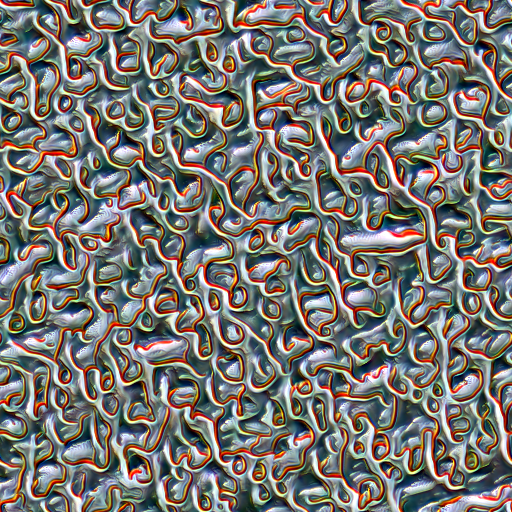}
    \end{subfigure}
    \begin{subfigure}{0.31\textwidth}
        \includegraphics[width=\textwidth]{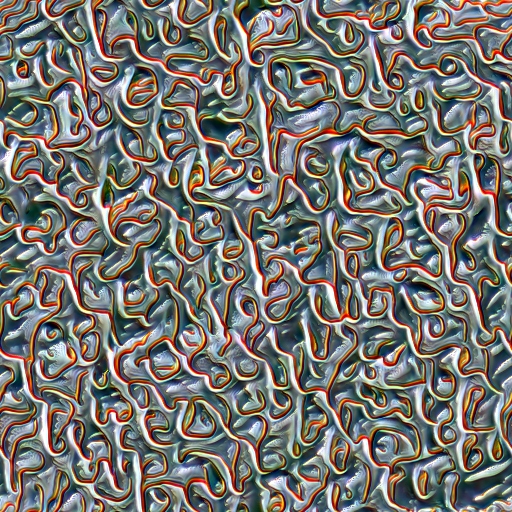}
    \end{subfigure}
    
    \begin{sideways} NCA \& Perlin Noise\end{sideways}\quad
    \begin{subfigure}{0.31\textwidth}
        \includegraphics[width=\textwidth]{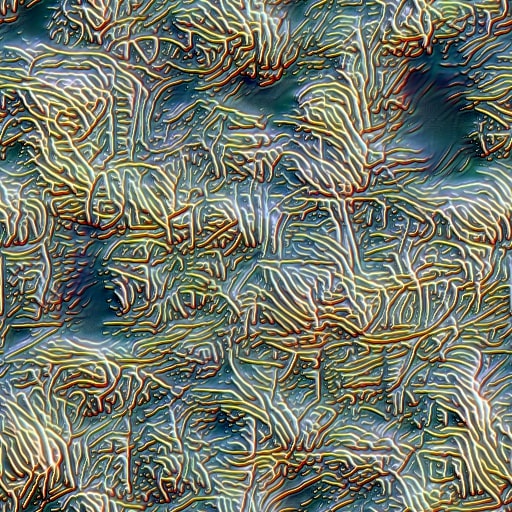}
    \end{subfigure}
    \begin{subfigure}{0.31\textwidth}
        \includegraphics[width=\textwidth]{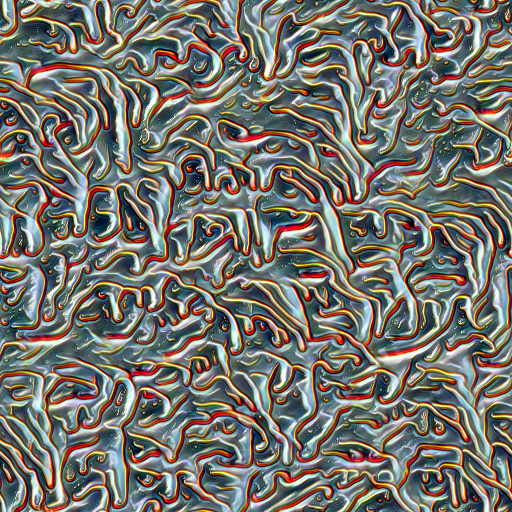}
    \end{subfigure}
    \begin{subfigure}{0.31\textwidth}
        \includegraphics[width=\textwidth]{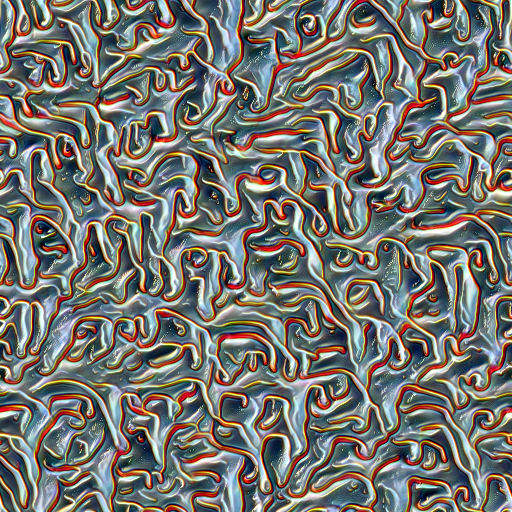}
    \end{subfigure}
    
    \begin{sideways} MNNCA \& Uniform Noise\end{sideways}\quad
    \begin{subfigure}{0.31\textwidth}
        \includegraphics[width=\textwidth]{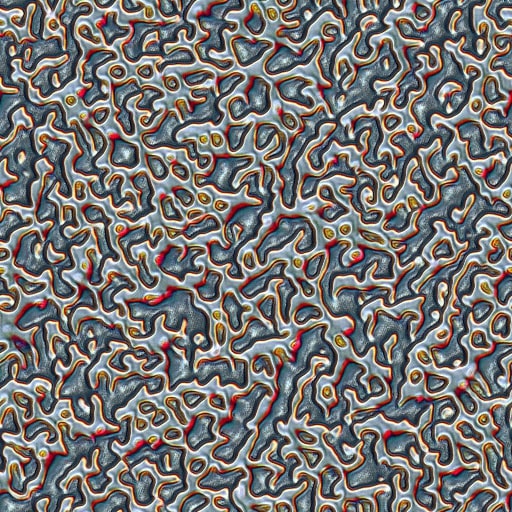}
    \end{subfigure}
    \begin{subfigure}{0.31\textwidth}
        \includegraphics[width=\textwidth]{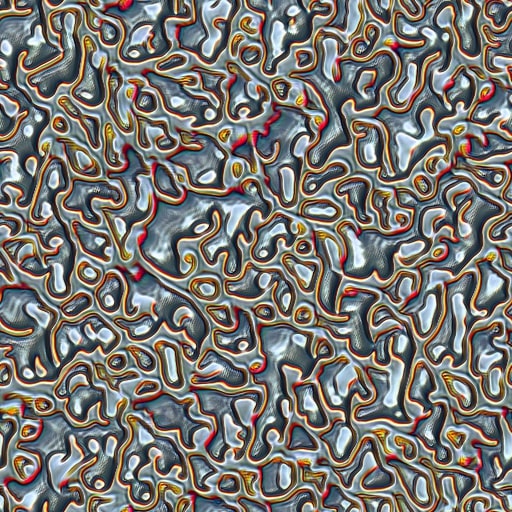}
    \end{subfigure}
    \begin{subfigure}{0.31\textwidth}
        \includegraphics[width=\textwidth]{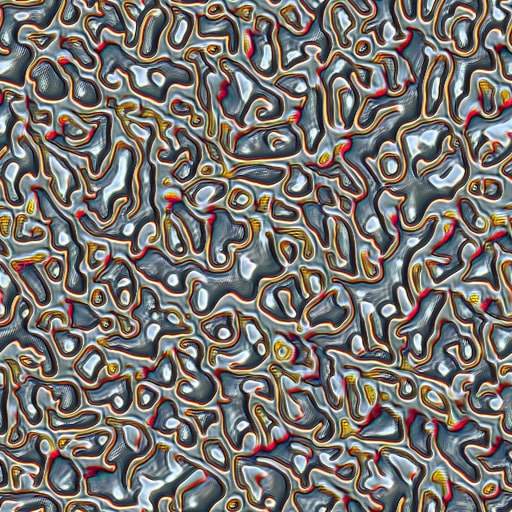}
    \end{subfigure}
    
    \begin{sideways} MNNCA \& Perlin Noise\end{sideways}\quad
    \begin{subfigure}{0.31\textwidth}
        \includegraphics[width=\textwidth]{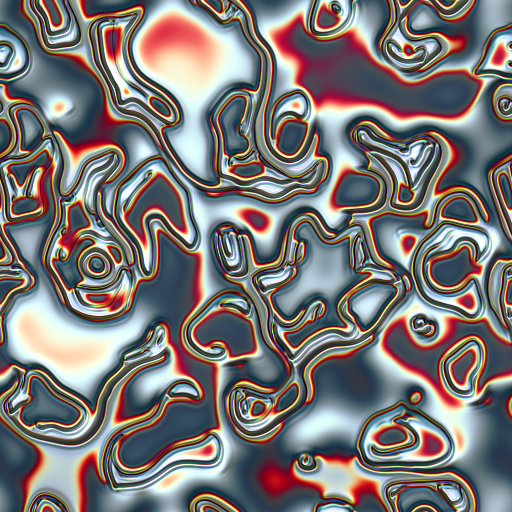}
    \end{subfigure}
    \begin{subfigure}{0.31\textwidth}
        \includegraphics[width=\textwidth]{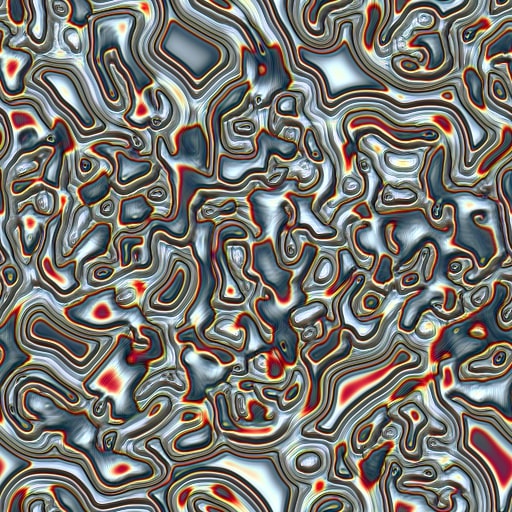}
    \end{subfigure}
    \begin{subfigure}{0.31\textwidth}
        \includegraphics[width=\textwidth]{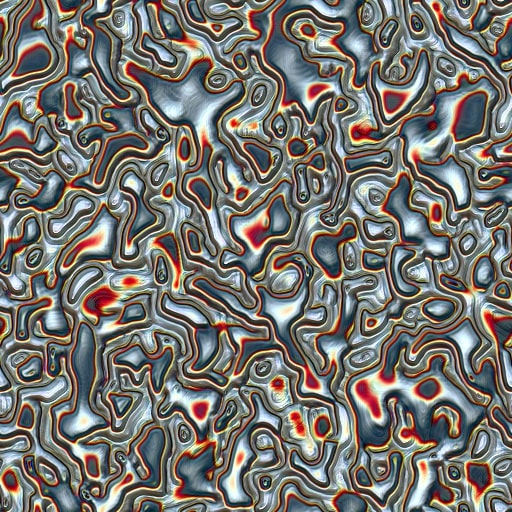}
    \end{subfigure}
    
    \caption{Illustrations of various timesteps, 50, 200, and 600, of classical NCA and MNNCA, with uniform and Perlin noise. \href{https://images-wixmp-ed30a86b8c4ca887773594c2.wixmp.com/f/056fb59c-3e38-4de4-813f-a3cc869f33e7/dees4mu-1b402149-a698-4cd3-b384-483ba01ebd9e.jpg/v1/fill/w_894,h_894,q_70,strp/stock__psychedelic_texture_by_fractalcaleidoscope_dees4mu-pre.jpg?token=eyJ0eXAiOiJKV1QiLCJhbGciOiJIUzI1NiJ9.eyJzdWIiOiJ1cm46YXBwOjdlMGQxODg5ODIyNjQzNzNhNWYwZDQxNWVhMGQyNmUwIiwiaXNzIjoidXJuOmFwcDo3ZTBkMTg4OTgyMjY0MzczYTVmMGQ0MTVlYTBkMjZlMCIsIm9iaiI6W1t7ImhlaWdodCI6Ijw9MTI4MCIsInBhdGgiOiJcL2ZcLzA1NmZiNTljLTNlMzgtNGRlNC04MTNmLWEzY2M4NjlmMzNlN1wvZGVlczRtdS0xYjQwMjE0OS1hNjk4LTRjZDMtYjM4NC00ODNiYTAxZWJkOWUuanBnIiwid2lkdGgiOiI8PTEyODAifV1dLCJhdWQiOlsidXJuOnNlcnZpY2U6aW1hZ2Uub3BlcmF0aW9ucyJdfQ.AxN2NxGw4V6MBnb-SlzlwEUC5TdVlLP4Njb6QS1XZk4}{Trained on texture} \cite{fractalcaleidoscope2023}.}
    \label{fig:algorithm_stages}
\end{figure}
\newpage
\subsubsection{Comparison with Fixed Parameter Count}
To ensure a fair comparison, given that MNNCAs inherently possess more parameters, we trained an NCA with a parameter size comparable to that of MNNCA. Details of this can be found in Section~\ref{sec:ExperimentalDetails}. The observed results suggest that the enhanced capability of MNNCAs to generate complex textures is not solely attributed to their increased parameter count as the NCA still generates less structured textures.
\begin{figure}[htbp]
    \centering
    
    \begin{sideways} NCA  \end{sideways}\quad
    \begin{subfigure}{0.31\textwidth}
        \includegraphics[width=\textwidth]{Figures/nca_basic_image_at_step_50.png}
    \end{subfigure}
    \begin{subfigure}{0.31\textwidth}
        \includegraphics[width=\textwidth]{Figures/nca_basic_image_at_step_200.png}
    \end{subfigure}
    \begin{subfigure}{0.31\textwidth}
        \includegraphics[width=\textwidth]{Figures/nca_basic_image_at_step_600.png}
    \end{subfigure}
    
    \begin{sideways} NCA XL \end{sideways}\quad
    \begin{subfigure}{0.31\textwidth}
        \includegraphics[width=\textwidth]{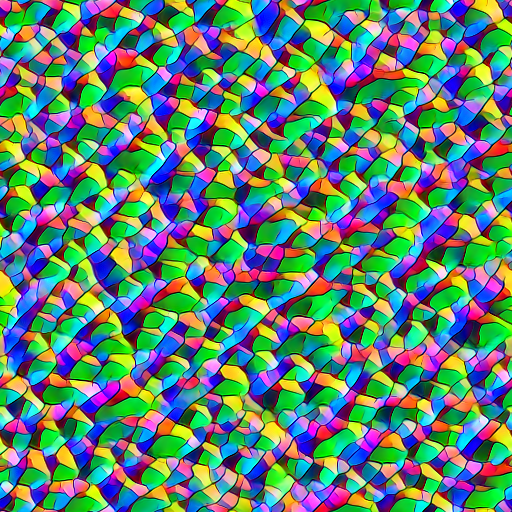}
    \end{subfigure}
    \begin{subfigure}{0.31\textwidth}
        \includegraphics[width=\textwidth]{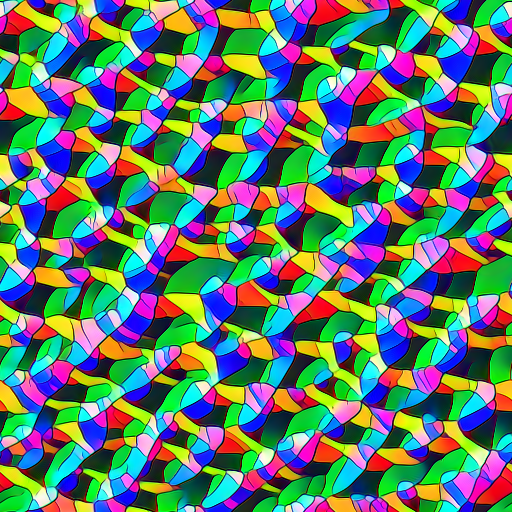}
    \end{subfigure}
    \begin{subfigure}{0.31\textwidth}
        \includegraphics[width=\textwidth]{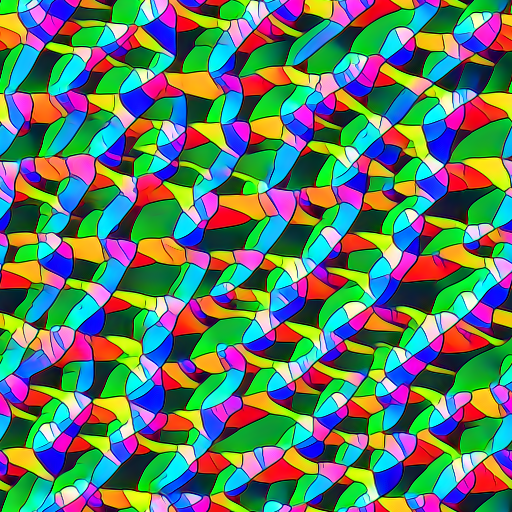}
    \end{subfigure}
    \caption{Illustrations of various timesteps, 50, 200, and 600, of classical NCA with uniform noise and different sizes.\href{https://i0.hippopx.com/photos/521/378/436/digital-multicolor-colorful-curves-preview.jpg}{Trained on texture} \cite{hippopx2023image}.}
    \label{fig:algorithm_stages}
\end{figure}
\end{document}